\pdfoutput=1

\documentclass[11pt]{article}
\usepackage[x11names]{xcolor} 

\usepackage{acl}

\usepackage{algorithm}
\usepackage{algorithmic}

\usepackage{times}
\usepackage{latexsym}
\usepackage{graphicx}
\usepackage{adjustbox}
\usepackage{changepage}
\usepackage{geometry}

\usepackage[utf8]{inputenc}

\usepackage{hyperref}       
\usepackage{url}            
\usepackage{booktabs}       
\usepackage{amsfonts}       
\usepackage{nicefrac}       
\usepackage{microtype}      

\usepackage{kotex}
\usepackage{verbatim}
\usepackage{multirow}
\usepackage{bbm}
\usepackage{mathtools}
\usepackage{caption}
\usepackage{subcaption}
\usepackage{graphicx}
\newcommand*{\affmark}[1][*]{\textsuperscript{#1}}

\title{PePe: Personalized Post-editing Model utilizing User-generated Post-edits}

\author{Jihyeon Lee\affmark[*]\\
{\tt\small gina.ai@kakaobrain.com}
\And
Taehee Kim\affmark[*]\\
{\tt\small taehee.kim@letsur.ai}
\And
Yunwon Tae\affmark[*]\\
{\tt\small yunwon.tae@vuno.co}
\AND
Cheonbok Park\\
{\tt\small cbok.park@navercorp.com}
\And
Jaegul Choo\\
{\tt\small jchoo@kaist.ac.kr}
}

\newcommand{\jh}[1]{\textcolor{black}{{#1}}}

\newcommand{\taehee}[1]{\textcolor{black}{{#1}}}
\newcommand{\commented}[1]{\textcolor{black}{{#1}}}

\begin{document}

\maketitle
\begin{abstract}
Incorporating personal preference is crucial in advanced machine translation tasks. 
Despite the recent advancement of machine translation, it remains a demanding task to properly reflect personal style.
In this paper, we introduce a personalized automatic post-editing framework to address this challenge, which effectively generates sentences considering distinct personal behaviors.
To build this framework, we first collect post-editing data that connotes the user preference from a live machine translation system.
Specifically, real-world users enter source sentences for translation and edit the machine-translated outputs according to the user's preferred style. 
We then propose a model that combines a discriminator module and user-specific parameters on the APE framework.
Experimental results show that the proposed method outperforms other baseline models on four different metrics (\textit{i.e.,} BLEU, TER, YiSi-1, and human evaluation). 
\end{abstract}

\section{Introduction}
Language usage is strongly influenced by the state of the individual, which can be considered by multiple attributes such as age, gender, socioeconomic status, and occupation~\citep{tannen1991you, pennebaker2003psychological}. 
Taking these aspects into account in the machine translation task, we need personalized translations to reflect individual characteristics that vary from person to person; thus, the translation system should consider not only fluency and content preservation, but also personal style. 

However, most existing neural machine translation (NMT) models ignore personal style~\citep{mirkin-etal-2015-motivating}. 
Previous studies attempt to address this problem by personalizing the NMT models, but in these studies the definition of personal style is often over-simplified. 
For example, \citet{rabinovich-etal-2017-personalized} and \citet{sennrich-etal-2016-controlling} define the personal style as politeness and gender respectively, which is not sufficient to tackle the multifarious character of an individual.
Namely, previous works defined the personal style in a constrained form.

\begin{figure}
\begin{center}
\includegraphics[width=\linewidth]{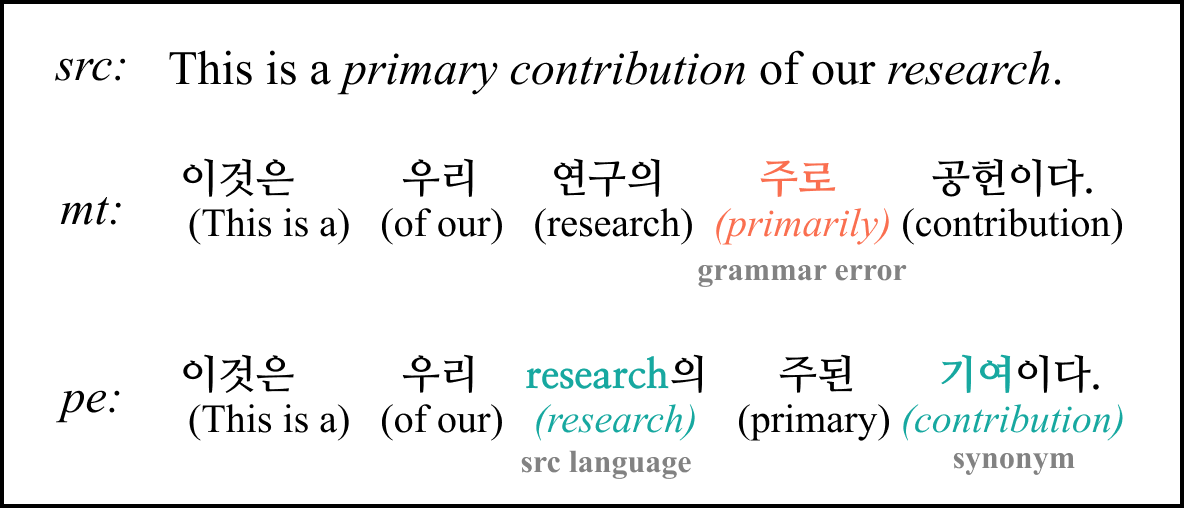}
\end{center}
   \caption{Example of a \textit{personal post-editing} triplet (i.e., source (src), machine translation (mt), and post-edit (pe)) given the source text in English and the translated text in Korean. A post-edited sentence does not only contain error correction of an initial machine translation result but also reflects individual preference. For instance, a human post-editor modifies the word "primarily" to "primary," but also change "공헌" to its synonym "기여" while keeping the rest as it is (e.g., "research").}
\label{fig:ape_example}
\end{figure}

In contrast with previous studies, we propose a method based on an APE framework and newly utilize post-editing data to capture diverse personal traits in translation.
Originally, the need for post-editing data is to improve the quality of machine-translated sentences in an APE task~\citep{simard-etal-2007-rule, pal-etal-2016-neural,correia-martins-2019-simple}. However, we suggest that the post-editing data can also be adequate references for personalized translation if various users post-edit sentences according to their preferences.
In this respect, we collect a \textbf{us}er-generated \textbf{p}ost-editing dataset called USP through a live translation system. 
After the system translates a source sentence (\textit{src}) to a target sentence, \textit{i.e.,} machine-translated sentence (\textit{mt}), each user edits the translated result according to their purpose or preferences, \textit{i.e.,} post-edited sentence (\textit{pe}). 
We collect (\textit{src, mt, pe}) triplets called \textit{personalized post-editing} triplets for each user and an example is depicted in Fig~\ref{fig:ape_example}. 

Along with the personalized post-editing data, we develop a model which utilizes user parameter and a discriminator module.
The user-specific parameters allow the model to adapt to each user in that the model can consider inter-personal variations. 
These parameters are aggregated with the output word probability such that the generation word probability distribution differs by each particular user. 
Moreover, since the prevalence of pre-trained language models encourages significant performance improvements on various natural language generation tasks~\citep{song2019mass, lewis2019bart, correia-martins-2019-simple}, we exploit the pre-trained language model (LM) but do not fully lean on it. We assume that not all the features from the pre-trained LM contribute to capturing the distinct taste of users that are departing from the neutral and standardized patterns.
Thus, our discriminator module, inspired by adversarial training~\citep{goodfellow2014generative}, attempts to dismantle the unnecessary features from a pre-trained LM, while tuning the model to incorporate a personal style.
The details will discuss in Section~\ref{method}.

Experiments on our dataset and speaker annotated TED talks dataset~\citep{michel-neubig-2018-extreme} (SATED) demonstrate that the proposed approach generates diverse translations for different users.

In summary, our contributions include the following:
\begin{itemize}
   \setlength\itemsep{0.1em}
    \item To the best of our knowledge, this is the first work that leverages the APE framework to a personalized translation task.

    \item We propose a personalized post-editing model based on user-generated post-edits, which is able to capture the inter-personal variations that consist of multiple attributes.

    \item Extensive experimental results show that the proposed method robustly reflects personal traits and consistently outperforms baselines in three different quantitative metrics and human evaluation results. 
\end{itemize}

\begin{figure*}
\begin{center}
\begin{adjustbox}{width=0.95\textwidth}
\includegraphics[width=\linewidth]{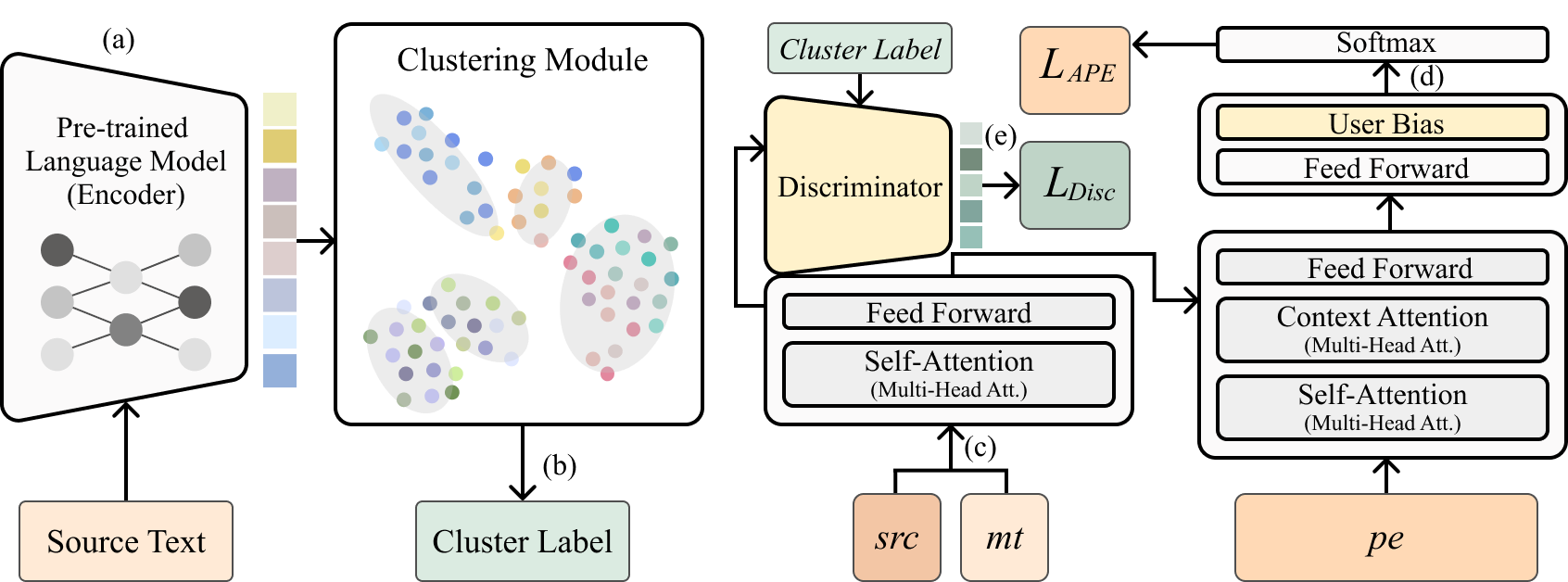}
\end{adjustbox}
\end{center}
\vspace{-0.2cm}
\caption{An overview of our proposed method. PePe consists of two parts: 1) Clustering module that relies on pre-trained LM encoder and Gaussian mixture model. 2) APE architecture that includes an auxiliary discriminator and user-specific parameters.
}
\vspace{-0.3cm}\label{fig:main_ah}
\end{figure*}

\section{Related Work}

Our work is closely related to the recent work on personalized neural machine translation and automatic post-editing. 

\paragraph{\textbf{Personalized neural machine translation.\phantom{aaaa}}}

Standard NMT systems are not able to consider the personal preference in a machine-translated output~\citep{mirkin-etal-2015-motivating}. 
\citet{mima1997improving} is the early paper that proposes a concept of reflecting an author's properties, such as gender, dialog domain, and role in the translation. 
However, including \citet{mima1997improving}, most studies conduct a limited range of personalized translations, which address only a single attribute (\textit{e.g.}, politeness)~\citep{sennrich-etal-2016-controlling, rabinovich-etal-2017-personalized}. 

\citet{turchi2017continuous} and \citet{karimova2018user} fine-tune the model on the human post-edits to improve the NMT quality, which can be viewed as a naive approach to handle the personalized translation without attribute labels.
\citet{wuebker-etal-2018-compact} extend this approach to adjust only a small number of parameters, but still requires extensive training resources.
Meanwhile, \citet{michel-neubig-2018-extreme} and \citet{user-driven} propose a generalized form of a personalized translation method, which are closely related work with ours.
\citet{michel-neubig-2018-extreme} cast this problem as an extreme form of domain adaptation, while \citet{user-driven} introduce cache module and contrastive learning to increase the diversity on dissimilar users.
However, the reference sentences for personalized translation were constructed by a few professional translators, not by a variety of people with diverse characteristics; personal preferences reflected in the dataset are limited.
Our user-generated post-edits are edited by a large number of people who provide the original sentences.

\paragraph{\textbf{Automatic post-editing.}}
Prior to the emergence of the transformer~\citep{vaswani2017attention}, RNN based APE models~\citep{pal-etal-2016-neural, junczys2016amu, junczys2017exploration} are actively studied. 
Subsequently, as self-attention based models show significant improvements on various downstream tasks, transformer based models also prevail in the APE task.
Specifically, a popular approach is to set a separate encoder for the source and machine-translated (MT) output. 
Separately encoded representations are joined in the following encoder~\citep{pal2018transformer} or fused in the decoder~\citep{tebbifakhr-etal-2018-multi, junczys2018ms}.
More recently, \citet{correia-martins-2019-simple} improve the performance of APE tasks by leveraging a pre-trained LM.
Compared to these studies, our work is the first attempt to examine the neural network based APE model for personalized translation. 
There is a study where they use an APE module for domain adaptation~\citep{isabelle2007domain}, but the explored one is based on a statistical machine translation system.

\begin{figure}
     \centering
     \begin{subfigure}[b]{0.45\columnwidth}
         \centering
         \includegraphics[width=\linewidth]{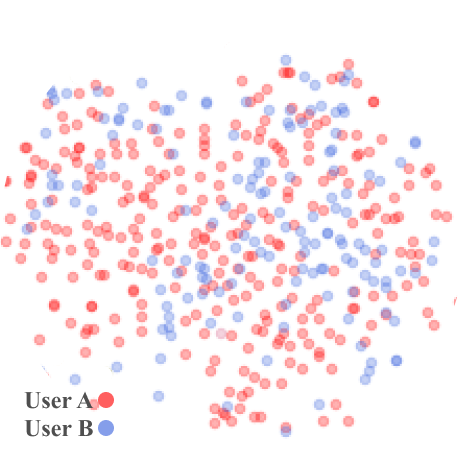}
         \caption{Data points colorized with user label}
         \label{fig:2a}
     \end{subfigure}
     \hfill
     \begin{subfigure}[b]{0.45\columnwidth}
         \centering
         \includegraphics[width=\linewidth]{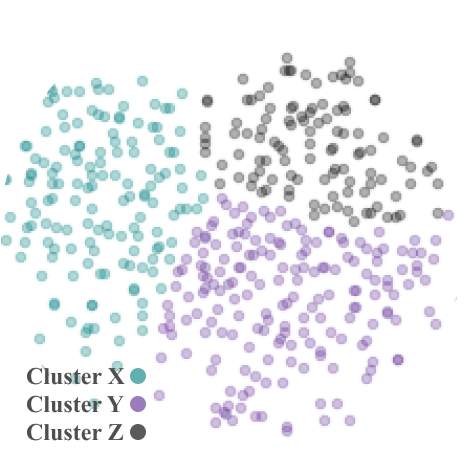}
         \caption{Data points colorized with GMM cluster label}
         \label{fig:2b}
     \end{subfigure}
         \caption{Source sentences of USP embedded with the pre-trained LM. (a) and (b) shows the discrepancy between the user data distribution and the contextual similarity-based data distribution.}
        \label{fig:2cluster}
\end{figure}

\section{Proposed Method}\label{method}
\paragraph{\textbf{Overview:}}
It is challenging to generate appropriate translations that impose personal variations. 
To address such a demanding problem, we take a detour by applying an APE framework.
We propose PePe, a personalized post-editing model utilizing user-generated post-edits.
\commented{PePe includes a discriminator module to allow the model to dismantle the pre-trained LM features.} Specifically, we maximize the discriminator loss to encourage the encoder to throw away irrelevant pre-trained LM features, while minimizing the APE loss to guide the model to utilize the pre-trained LM features that are useful for personalization. \commented{In addition, PePe utilizes user-specific parameters to capture the personal style.} User-specific parameters are combined at the end of the decoder layer to adjust the prediction of the word probability, i.e., the word choice based on a user preference. Our strategy does not require expensive supervision on the personal style, such as explicit attribute labeling or an attribute-tailored model architecture.

The overall architecture of PePe is illustrated in Fig.~\ref{fig:main_ah}.
The two following subsections will describe the modules shown in Fig.~\ref{fig:main_ah}-(a), (b), (c), (d), and (e), respectively.

\subsection{Contextual Similarity vs. User-specific Style}\label{unnecessary}
\commented{The pre-trained LM is well known for capturing the contextual similarity that is useful to define the label for in-domain data (e.g., sports, IT, and economy).
However, the user-specific style is far from those domains; it does not coincide with contextual similarity yet involves somewhat arbitrary traits (\textit{i.e.,} user preferences).} 
Hence, we argue that some of the features from a pre-trained LM distract personalized translation, which rather requires generating biased results to meet the individuals' needs. 
Fig.~\ref{fig:2cluster} demonstrates the discrepancy between the user data distribution and the contextual similarity-based data distribution.




We map the sampled sentences from USP to the embedding space of the pre-trained LM.
Each sentence is encoded with RoBERTa~\citep{liu2019roberta} and visualized using t-SNE~\cite{van2008visualizing}. The data on both sides show the same embedding representations obtained from the same set of sentences, but labeling is different.
The data items in Fig.~\ref{fig:2a} are color-coded by the users, whereas those in Fig.~\ref{fig:2b} are color-coded by the semantic cluster labels obtained from the Gaussian mixture model (GMM)~\citep{rasmussen2000infinite}, which allocates the similar sentences to the same label based on the RoBERTa embedding of each.

In Fig.~\ref{fig:2b}, semantically similar points, which are close in embedding space, belong to the same clusters.
However, the red and blue points in Fig.~\ref{fig:2a}, which indicate sentence representations from two different users, are distributed unruly instead of being grouped by semantic similarity.
In other words, the fine-grained style differences of each user are somewhat distant from the contextual similarity of the sentences; thus it is hard to distinguish user-specific preferences when the model is highly oriented to learning the contextual similarity.

\subsection{Generating Cluster Labels based on Pre-trained LM}
\commented{Inspired by the finding in Section~\ref{unnecessary}, we devise a discriminator module that uses the semantic cluster labels to unlearn the features from the pre-trained LM that are unnecessary to reflect the personal styles.}
Before introducing the details about PePe, we describe how to generate the semantic cluster labels from a pre-trained LM in an unsupervised manner.
We first encode \textit{src} into encoded vectors using a pre-trained LM\footnote{Though we use RoBERTa as a pre-trained LM to generate cluster labels, other pre-trained LMs can also be used in our approach.} as shown in Fig~\ref{fig:main_ah}-(a).
Based on these encoded vectors, semantic cluster labels are generated by GMM~\citep{rasmussen2000infinite} as illustrated in Fig~\ref{fig:main_ah}-(b).
A Gaussian mixture is a function made up of the $k$ number of Gaussian components, where $k$ is the number of clusters\footnote{Ten clusters are used for all the experiments in the main paper.} and is a hyperparameter. 
Specifically, in GMM, $\sum^{k}_{i=1}\pi_{i}p_{i}(\mathbf{h}|\theta_{i})$ represents the distribution of data point $\mathbf{h}$, where $\mathbf{h}$ is an encoded vector of the first token of \textit{src}, \textit{i.e.,} [CLS] token,
$\pi_{i}$ is the probability of each Gaussian fitting the data, and $p_{i}$ is the Gaussian density function parameterized by $\theta_{i}$.
We assign each sentence to a Gaussian that best describes the data, and the Gaussian corresponds to the semantic cluster label. The label, \textit{i.e.,} $T={t_{1},...,t_{k}}$, is then used as a classification label for our discriminator, which is described in the following subsection.

\subsection{PePe: Personalized Post-editing Model utilizing User-generated Post-edits}
\label{method:adversarial_learning}
We adopt BERT-based Encoder-Decoder APE model~\citep{correia-martins-2019-simple} called Dual-Source BERT (DS-BERT) as our backbone, which is based on transformer~\citep{vaswani2017attention} with pre-trained multilingual BERT~\citep{devlin-etal-2019-bert}. 
This approach is shown to outperform existing APE models~\citep{tebbifakhr-etal-2018-multi, junczys-dowmunt-grundkiewicz-2018-ms, huang-etal-2019-learning-copy}. DS-BERT uses a single encoder which is used to encode both the \textit{src} and the \textit{mt} by concatenating them with the specialized token $[SEP]$ as described in Fig~\ref{fig:main_ah}-(c).


Our model also learns to generate $y=[y_{1},...,y_{n}]$, \textit{i.e.,} \textit{pe}, 
from $x$, \textit{i.e.,} \textit{src}, and $\Tilde{y}=[\Tilde{y}_{1},...,\Tilde{y}_{m}]$, \textit{i.e., mt}, by maximizing the likelihood, 
\begin{equation*}\label{eqn:ape}
    P(y|x,\Tilde{y};\theta_{\textit{APE}}) = \prod^{n}_{i=1}P(y_{i}|x,\Tilde{y},y_{<i};\theta_{\textit{APE}}),
\end{equation*}
where $y_{i}$ is the \textit{i}-th target word and $y_{<i}=y_{1}...y_{i-1}$ is the partial translation result. $\theta_{\textit{APE}}$ represents the parameters for translating source sentence into post-edited sentence with machine-translated result $\Tilde{y}$.





In order to adapt user-specific linguistic styles, we add user-specific parameters before the softmax layer in the decoder as shown in Fig~\ref{fig:main_ah}-(d), \textit{i.e.,} 
\begin{equation*}\label{eqn:ape_loss}
    P(y_{i}|x,\Tilde{y},y_{<i};\theta_\textit{APE}, \theta_{user}) = f(FFN(o_{i}) + \theta_{user}),
\end{equation*}
where \textit{FFN} and $f$ are a feed-forward network and softmax function, respectively. $o_{i}$ is the output for the \textit{i}-th target word from the decoder. $\theta_{user} \in \mathbb{R}^{V}$ is a user-specific embedding vector from a set of trainable user embedding matrix $U \in \mathbb{R}^{N\times{V}}$ where $N$ is the number of users and $V$ is the size of vocabulary.

The model is then optimized by minimizing $\mathcal{L}_{APE}$ defined as
\begin{equation*}
    \mathcal{L}_{\textit{APE}} = -\sum^{n}_{i=1}logP(y_{i}|x,\Tilde{y};\theta_{\textit{APE}},\theta_{user}).
\end{equation*}
Furthermore, as shown in Fig~\ref{fig:main_ah}-(e), we introduce a discriminator module to unlearn the contextual similarity feature learned from a pre-trained LM.
To train the discriminator, we compute the discriminator loss $\mathcal{L}_{\textit{Disc}}$ defined as
\begin{equation*}
    \mathcal{L}_{\textit{Disc}} = \sum^{k}_{i} t_{i}log(\Tilde{t}_{i}),
\end{equation*}
where $k$ is the number of classes (\textit{i.e.,} the number of Gaussians we pre-defined) and $t_{i}$ is the ground-truth label of the semantic cluster.
$\Tilde{t}_{i}$ represents the output from the discriminator which is a single-layer feed-forward network for the classification of semantic cluster labels.
We use the first token of a source sentence to extract a sentence representation from the encoder and pass it to the discriminator as an input. 
Note that we use the gradient ascent method to prevent the encoder from classifying the clusters. \commented{In this way, we diminish the unnecessary feature from pre-trained LM, while our APE loss function incorporated with user-specific parameters leads the model to capture the user-specific style.}

Finally, PePe optimizes a combination of two losses, $\mathcal{L}_{\textit{Disc}}$ and $\mathcal{L}_{\textit{APE}}$, with a adjustment rate $\mathcal{\beta}$, \textit{i.e.,} 
\begin{equation*}
    \mathcal{L}_{\textit{Train}}=\mathcal{\beta} \cdot \mathcal{L}_{\textit{Disc}} + (1-\mathcal{\beta}) \cdot \mathcal{L}_{\textit{APE}}.
\end{equation*}

\begin{table*}
\begin{center}
\begin{adjustbox}{width=1\textwidth}
\begin{tabular}{lcccccccc}
\toprule
\multicolumn{1}{l}{\multirow{2}{*}{Methods}} & 
\multicolumn{3}{c}{\textit{en$\rightarrow$ko}} & 
\multicolumn{1}{c}{} & 
\multicolumn{3}{c}{\textit{ko$\rightarrow$en}} \\
\cmidrule{2-4} \cmidrule{6-8}
& \multicolumn{1}{c}{BLEU$\uparrow$} & \multicolumn{1}{c}{TER$\downarrow$} & \multicolumn{1}{c}{YiSi-1$\uparrow$} &&
\multicolumn{1}{c}{BLEU$\uparrow$} & \multicolumn{1}{c}{TER$\downarrow$} & \multicolumn{1}{c}{YiSi-1$\uparrow$} \\
\midrule
(1) Uncorrected     & 64.9~\small{(-5.6)}   & 21.1~\small{(+1.2)}   & 87.3~\small{(-1.1)}   && 75.1~\small{(-3.5)}   & 17.7~\small{(+1.4)}   & 88.9~\small{(-0.8)} \\ \midrule
(2) DS-BERT & 68.4~\small{(-2.1)}   & 21.1~\small{(+1.2)}   & 87.6~\small{(-0.8)}   && 77.1~\small{(-1.5)}   & 17.6 ~\small{(+1.3)}   & 89.1~\small{(-0.6)} \\
(3) DS-BERT + Full Bias & 68.6~\small{(-1.9)}   & 20.9~\small{(+1.0)}   & 88.0~\small{(-0.4)}   && 78.0~\small{(-0.6)}   & 16.9 ~\small{(+0.6)}   & 89.6$^\ast$~\small{(-0.1)} \\
(4) DS-BERT + Factor Cell & 67.5~\small{(-3.0)}   & 22.1~\small{(+2.2)}   & 88.0~\small{(-0.4)}   && 76.5~\small{(-2.1)}   & 18.4 ~\small{(+2.1)}   & 89.2~\small{(-0.5)} \\
(5) DS-BERT + User CLS & 69.0~\small{(-1.5)}   & 20.9~\small{(+1.0)}   & 87.1~\small{(-1.3)}   && 78.1~\small{(-0.5)}   & 16.5 ~\small{(+0.2)}   & 89.4~\small{(-0.3)} \\
(6) DS-BERT + User Token & 68.8~\small{(-1.7)}   & 20.5~\small{(+0.6)}   & 87.0~\small{(-1.4)}   && 74.3~\small{(-4.3)}   & 21.6 ~\small{(+5.3)}   & 88.5~\small{(-1.2)} \\ \midrule
(7) \textbf{PePe}  & \textbf{70.5} & \textbf{19.9} & \textbf{88.4} && \textbf{78.6} & \textbf{16.3} & \textbf{89.7}\\ \midrule
(8) -discriminator & 68.6~\small{(-1.9)}  & 20.9~\small{(+1.0)}  & 88.0~\small{(-0.4)} && 78.0~\small{(-0.6)}  & 16.9~\small{(+0.6)} & 89.6$^\ast$~\small{(-0.1)}\\
(9) -(8) \& user bias & 68.4~\small{(-2.1)}    & 21.1~\small{(+1.2)}  & 87.6~\small{(-0.8)} && 77.1~\small{(-1.5)}  & 17.6~\small{(+1.3)} & 89.1~\small{(-0.6)}\\
(10) -(9) \& pre-training & 60.2~\small{(-10.3)} & 31.9~\small{(+12.0)} & 86.3~\small{(-2.1)} && 67.6~\small{(-11.0)} & 28.7~\small{(+12.4)} & 87.6~\small{(-2.1)}\\
\bottomrule
\end{tabular}
\end{adjustbox}
\end{center}
\vspace{-0.2cm}
\caption{Quantitative comparison with the baselines on the USP dataset that contains \textit{en$\rightarrow$ko} language pairs and vice versa. (8) to (10) denotes the ablation results. 
The ablation study is designed to verify each module in PePe.
User bias in (9) denotes the user-specific parameters located at the end of the decoder, and pre-training in (10) denotes the task-adaptive pre-training stage. 
The bold represents the significant difference ($p < 0.05$) against other baselines. We conduct the t-test with five runs and report the average score of it.
$^\ast$ means that there is no significant difference in the scores between the model and PePe.}
\vspace{-0.3cm}
\label{tab:main_result}
\end{table*}

\section{Experiments}
In this section, we qualitatively and quantitatively demonstrate the effectiveness of our proposed method.
We validate PePe, described in Section~\ref{method}, against other baseline methods using a real-world user dataset USP. 
We also provide a detailed explanation for the dataset.
Moreover, through extensive experiments and analyses, we show that PePe can incorporate \jh{inter-personal variations} into a target sentence.
We provide training details in Appendix~\ref{app: training details}.

\subsection{Dataset}\label{dataset}
\begin{figure}
\begin{center}
\includegraphics[width=\linewidth]{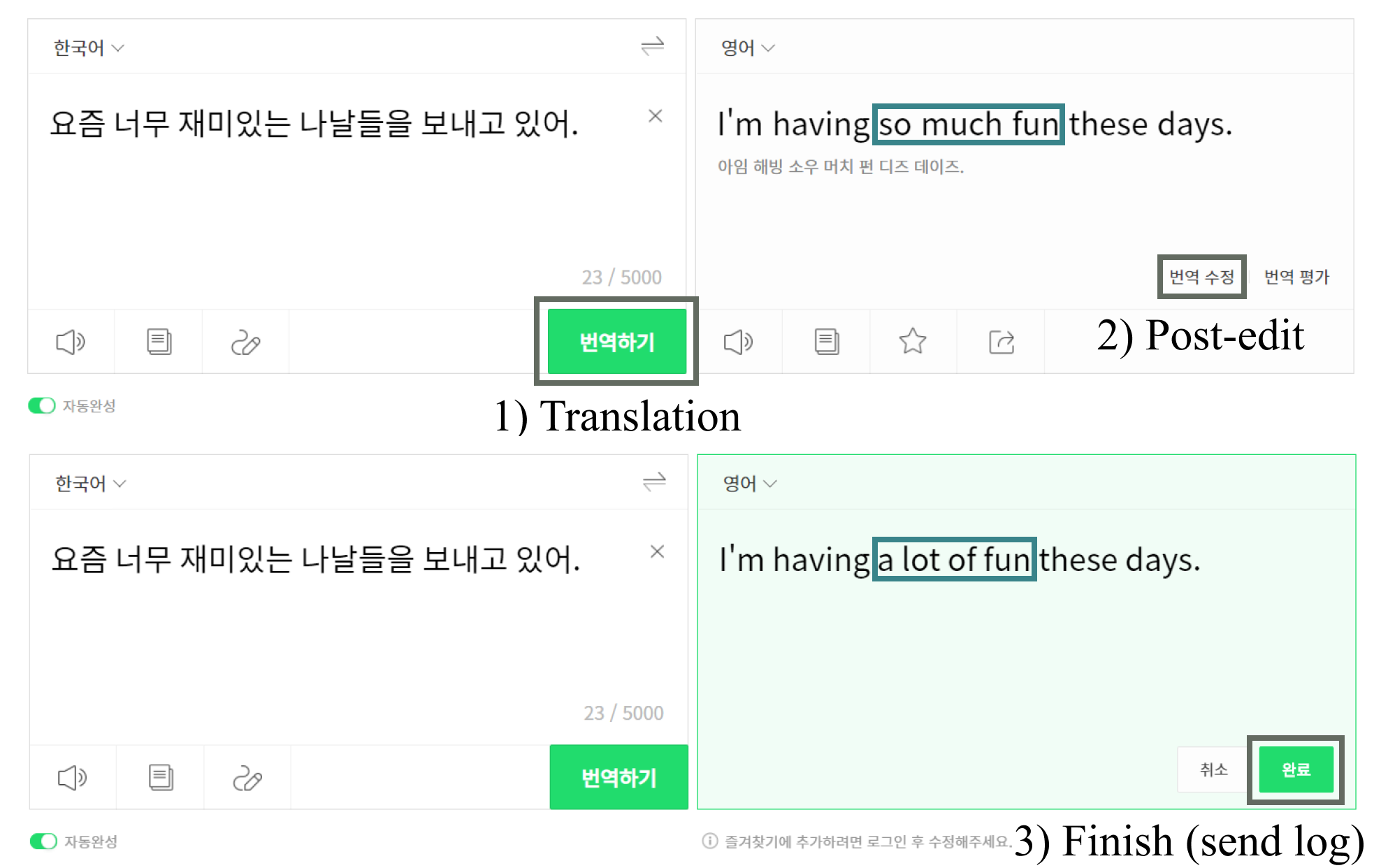}
\end{center}
  \caption{Illustration of the user experience flow for post-edit log generation.}
\label{fig:papago}
\end{figure}

We collect the user-generated post-editing dataset, USP, from a publicly available online translation system\footnote{We collected data only from users who consent to the data collection for research purposes. In addition, there is no privacy issue because de-identification had taken for the collected data.} (\textit{e.g.,} Google Translate). 
\jh{Fig~\ref{fig:papago}} illustrates the user experience flow.
The users enter the sentences they want to translate, and the system provides the corresponding outputs that are generated by the high-quality commercialized machine translator. 
From the machine-translated outputs, users can start to edit the translated sentences according to their preference by clicking the ``post-edit'' button.
Consequently, when the users click the ``Finish'' button after completing the changes, a triplet of the source sentence, machine-translated output, and personalized post-edit is sent to our database.
\jh{Note that the origin of post-edited sentences is each particular user, which makes USP contains inter-personal variation, unlike existing APE datasets.}

Since we collect USP from the real-world users' inputs that contain various noises (\textit{e.g.,} unedited, duplicated, or meaningless examples), we preprocess the data to eliminate these noises.
Furthermore, most users only edited few examples, which are not sufficient to represent their style. Therefore, we select the users who left more than 100 samples, \textit{i.e.,} 30 users with 7K sentences and 70 users with 9K sentences for \textit{en$\rightarrow$ko} and \textit{ko$\rightarrow$en} USP dataset, respectively.
For users who left less than 100 samples, we aggregate the samples (\textit{i.e.,} 0.12M sentences) and utilize them as training data for the task-adaptive pre-training ~\citep{gururangan2020don}.
The discriminator module and user-specific parameters are not used in the task-adaptive pre-training and only the parameters for DS-BERT are utilized for the pre-training stage.
Details of data preprocessing are in Appendix~\ref{app: training details}.

Additionally, we adopt a Speaker Annotated TED (SATED) dataset~\citep{michel-neubig-2018-extreme} containing more than 2,000 sets of speaker style-contained source sentences,
which is publicly available. We select the dataset to show the robustness of our model regarding different datasets and languages.

\subsection{Experimental Setup}\label{experimental}
\paragraph{\textbf{Evaluation metric.}}
We use three different metrics to evaluate how well our proposed model preserves the content and incorporates the personal \jh{preferences}. BLEU~\citep{papineni2002bleu} and TER~\citep{snover2006study} scores are considered to assess the translated sentence where the ground-truth sentence is a \textit{pe} sentence.
We also leverage YiSi-1~\citep{lo2019yisi} that computes the semantic similarity of phrases between the model output and \textit{pe}, which can be sensitive to detailed styles.
We also conduct a human evaluation, which will be described in the following section.

\paragraph{\textbf{Baseline Methods.}}

We compare the performance of our method with the following baselines. 
Since this is the first attempt to personalize the translation using post-edits, we newly adjust existing personalized translation methods onto the APE framework for comparisons.

\textbf{1) Uncorrected} is the same as \textit{mt} in personalized post-editing data, which is generated from the online MT system. No correction was made on it. 
\textbf{2) DS-BERT} is a transformer based post-editing model~\citep{correia-martins-2019-simple} that
we adopt as our backbone in the method section.
This approach is shown to outperform existing APE models~\citep{tebbifakhr-etal-2018-multi, junczys-dowmunt-grundkiewicz-2018-ms, huang-etal-2019-learning-copy}, even in the absence of additional large-scale artificial data that competing models have used. 
\textbf{3) DS-BERT + Full bias}~\citep{michel-neubig-2018-extreme} utilizes additional user bias vectors on the decoder's output.
\textbf{4) DS-BERT + Factor bias}~\citep{michel-neubig-2018-extreme} uses factorized user bias on the output of the decoder. User-independent biases are shared with all users. However, the user-specific vector can adjust each user-independent vector's magnitude.
\textbf{5) DS-BERT + User CLS} is a multi-task composed of a user classification and APE task. The first token of an encoder input is used to stand for user identity. 
The corresponding output vector is used to classify a ground-truth user label.
A single layer of a feed-forward neural network is used for the classifier.
\textbf{6) DS-BERT + User Token}~\citep{sennrich-etal-2016-controlling} adds a token at the start of each post-edited sentence to indicate the user for each sentence. We train the model in a teacher-forcing manner.

\subsection{Quantitative Evaluation}\label{quantitative}
Results using automatic metrics and human evaluation are presented in this section.
PePe consistently outperforms the baselines on all datasets we considered. We also show the robustness of PePe regarding the different number of users, data distributions, and language pairs.



\paragraph{\textbf{Performance of PePe against other baselines.}}
(1) to (7) in Table~\ref{tab:main_result} shows the personalized translation results of varied baselines.
Our proposed method outperforms the six baselines with the non-trivial margin both on \textit{en$\rightarrow$ko} and \textit{ko$\rightarrow$en} USP dataset.
For instance, BLEU score increased in the range of 1.7 to 5.6, YiSi-1 increased in the range of 0.4 to 1.4, and TER decreased in the range of 0.6 to 2.2 over baselines, in \textit{en$\rightarrow$ko} dataset.
Consistent results from these three different metrics verify that PePe easily \jh{figure out distinct taste of users} while preserving source contents.
Especially, experiments in \textit{en$\rightarrow$ko} dataset show the most outstanding performance gains since the data mostly come from the users whose first language is Korean; the users can reflect the linguistic preference more naturally on this dataset.

\paragraph{\textbf{Ablation study.}}
The comparison between PePe and (8) in Table~\ref{tab:main_result} shows the importance of the discriminator module. When we exclude the discriminator module, the BLEU and TER scores are decreased on both \textit{en$\rightarrow$ko} and \textit{ko$\rightarrow$en}.
The results of the vanilla APE model (\textit{i.e.,} (9) in Table~\ref{tab:main_result}) show that the user-specific parameters are also significant for personalized translation. 
Moreover, when we do not adopt the APE task-adaptive pre-training (\textit{i.e.,} (10) in Table~\ref{tab:main_result}), the performance of the model drops even further. Overall, our ablation study demonstrates that each component is essential for the task.

\begin{table}
\centering
\begin{tabular}{lccc}
\toprule
Metrics & PePe & DS-BERT & Uncorr.\\
\midrule
\small{\textbf{Style - 1st}} & \textbf{59.6} & 18.1 & 22.2 \\   
\small{\textbf{Style - 2nd}} & \textbf{21.0} & 39.1 & 39.9 \\
\small{\textbf{Style - 3rd}} & \textbf{19.5} & 42.6 & 37.9\\
\midrule
\small{\textbf{Non-Style}} & \textbf{3.94\small{(1.08)}} & 3.60\small{(1.19)} & 3.82\small{(1.16)}\\
\bottomrule
\end{tabular}
\caption{Human evaluation on \textit{en$\rightarrow$ko} USP dataset. Style and non-style factors are both surveyed. For the style factor, each score represents the proportion. For instance, 59.6\% of evaluators choose PePe as the first place among other models. For the non-style factor, a Likert scale from 1 to 5 evaluates fluency and source contents preservation. We report the average score and the standard deviation.}
\label{tab:human}
\end{table}

\begin{table}
\centering
\begin{adjustbox}{width=0.48\textwidth}
\begin{tabular}{lccc}
\toprule
\multirow{2}[1]{*}{Model}&\multicolumn{1}{c}
{\textit{en$\rightarrow$de}}&\multicolumn{1}{c}{}&
\multicolumn{1}{c}{\textit{en$\rightarrow$fr}}\\
\cmidrule{2-2} \cmidrule{4-4}
 & \small{BLEU$\uparrow$} && \small{BLEU$\uparrow$} \\ 
\midrule
\citet{michel-neubig-2018-extreme} & 27.2 && 38.5 \\
DS-BERT & 30.4 && 42.2 \\
\textbf{PePe} & \textbf{31.2} &&\textbf{43.7} \\
\bottomrule
\end{tabular}
\end{adjustbox}
\caption{\jh{Experiments on the SATED dataset. PePe outperforms DS-BERT on different language pairs even for a synthetic post-editing dataset. The bold represents the best score among the baselines and significantly ($p < 0.05$) outperforms DS-BERT.}}
\label{tab:ted}
\end{table}

\paragraph{\textbf{Human evaluation.}}
To validate the advantage of our approach, we conduct human evaluations. 
Human evaluation can be a reasonable measurement choice to evaluate the personalization task because even sophisticated evaluation metrics can fail to capture the abstract (\textit{i.e.,} high-level) user behavior reflected in the \textit{pe} sentence.
We hired 20 Korean-English who are bilingual and engaged in the fields of linguistics and machine learning for human evaluation. 
We randomly select 30 source sentences and generate corresponding target sentences from Uncorrected, DS-BERT, and PePe before carrying out two types of questions to compare different metrics. 
1) We ask participants to annotate generated sentences along with fluency and content preservation. Sentences are measured on a Likert scale from 1 to 5.
2) We take three sentences generated from three different models. Participants rank these sentences from first to third, \textit{i.e.,} asking which sentence is most similar to the ground-truth \textit{pe} that \jh{contains distinct writing styles.}

As reported in Table~\ref{tab:human}, PePe is ranked 1st by most evaluators. PePe not only achieved the best score on style evaluation but also on non-style factors (\textit{i.e.,} fluency and contents preservation), which is essential for the translation task.
DS-BERT achieves the lowest score on both measures, indicating that the ambiguous reflection of style is worse than none. 
The human evaluation results are consistent with our quantitative results measured by automatic metrics.

\begin{figure}
\begin{center}
\includegraphics[width=\linewidth]{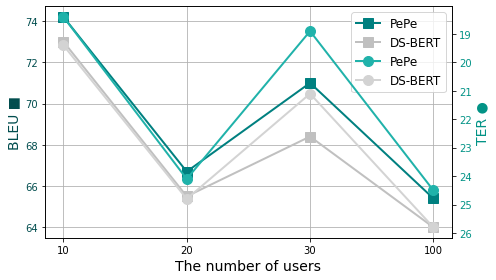}
\end{center}
    \vspace{-0.5cm}
   \caption{Robustness on the number of users. The dark squares denote the BLEU score, and the light circles denote the TER score. The number of clusters is equally adopted as ten for all cases. \jh{The users are randomly selected from the USP dataset.}}
\label{fig:robust}
\end{figure}

\definecolor{myr}{RGB}{255, 77, 106}
\definecolor{myg}{RGB}{23, 168, 160}
\definecolor{myo}{RGB}{252, 163, 98}

\begin{table*}
\begin{center}
\small
\begin{adjustbox}{width=0.84\textwidth}
\begin{tabular}{ll}
\toprule
\multicolumn{2}{l}{\textit{en$\rightarrow$ko}}\\
\midrule
\textit{src} & \textit{Immediately} provide non-monetary \underline{benefits} \underline{as required}. \\
\textit{mt} & \textcolor{myr}{필요에 따라} \textcolor{myo}{즉시} 비화폐성 \textcolor{myg}{편익}을 제공하십시오.  \\
\textbf{PePe} & \textcolor{myr}{\textbf{요구}된대로} 비화폐성 \textcolor{myg}{\textbf{혜택}}을 \textcolor{myo}{\textbf{즉시}} 제공한다. \\
\textit{pe} & \textcolor{myr}{\textbf{요구되어진}} 비화폐성 \textcolor{myg}{\textbf{혜택}}을 \textcolor{myo}{\textbf{즉시}} 제공한다.  \\
\midrule
\textit{src} & Choose this option to make the current \underline{preset load} whenever a new \underline{multi Instrument} is created. \\
\textit{mt} & 새 \textcolor{myr}{멀티 계측기}가 생성될 때마다 현재 \textcolor{myg}{사전 설정된 로드}를 만들려면 이 옵션을 선택하십시오.\\
\textbf{PePe} & 새 \textcolor{myr}{\textbf{multi instrument}}가 생성될 때마다 현재 \textcolor{myg}{\textbf{preset}}을 만들려면 이 옵션을 선택하십시오.\\
\textit{pe} & 새 \textcolor{myr}{\textbf{multi instrument}}가 생성될 때마다 현재 \textcolor{myg}{\textbf{preset load}}를 만들려면 이 옵션을 선택하십시오.\\
\midrule
\multicolumn{2}{l}{\textit{ko$\rightarrow$en}}\\
\midrule
\textit{src} & \underline{겨울왕국 2} 속에 나오는 장면이 있다.\\
\textit{mt} & there is a scene in \textcolor{myo}{winter kingdom 2}. \\
\textbf{PePe} & there is a scene in \textcolor{myo}{\textbf{frozen 2}.}\\
\textit{pe} & there is a scene in \textcolor{myo}{\textbf{frozen 2}.}\\
\midrule
\textit{src} & \underline{관사} 아래에 있는 모음코드가 이렇게 바뀌어진다.\\
\textit{mt} & the vowel code under the \textcolor{myg}{official building} changes like this.\\
\textbf{PePe} & the vowel code under the \textcolor{myg}{\textbf{article}} changes like this.\\
\textit{pe} & the vowel code under the \textcolor{myg}{\textbf{article}} changes like this.\\
\bottomrule
\end{tabular}
\end{adjustbox}
\end{center}
\vspace{-0.2cm}
\caption{Qualitative examples of post-edited sentences generated from PePe. User-specific parts in \textit{pe} and corresponding parts in \textit{mt} are colored. We highlight the post-edited words in PePe with bold if the words are identical to \textit{pe}. Corresponding parts in \textit{src} are underlined. Our model finds an appropriate combination of attributes in accordance with sentences and users.}
\vspace{-0.3cm}
\label{tab:qualitative analysis_main}
\end{table*}

\paragraph{\textbf{Robustness of our model.}}
Table~\ref{tab:ted} shows the personalized translation results on \textit{en$\rightarrow$de} and \textit{en}$\rightarrow$\textit{fr} SATED dataset.
Since the dataset is initially constructed for the machine translation task where post-edited sentences do not exist, we utilize target sentence (\textit{i.e., mt}) in the place of \textit{pe} and independently generate \textit{mt} from a particular translation model (\textit{i.e.,} pre-trained transformer based NMT model). 
Regardless of the language, the results demonstrate that PePe and DS-BERT, which leverages triplets (\textit{src}, \textit{mt}, \textit{pe}), \taehee{outperform} \citet{michel-neubig-2018-extreme} that relies on paired sentences (\textit{src}, \textit{mt}). In addition, the results also show that even if \textit{pe} is not edited from the \textit{mt}, PePe translates the source sentence close to the ground-truth target sentence that connotes the speaker's characteristics.

Fig.~\ref{fig:robust} shows that our model works well regardless of the number of users.
Grey-colored lines are the performance of the baseline model, and colored lines are the performance of PePe.
TER axis is reversed on the graph to make consistency with the BLEU score.
Note that the higher points denote a better score than the lower points.

Furthermore, we conduct additional experiments that show the robustness of our approach regarding the number of clusters and adjustment rates, which are hyperparameters. We represent the results in Appendix~\ref{app: robustness and hyperparameter}.

\subsection{Qualitative Analysis}
\commented{To understand how user-specific preferences are incorporated into the sentences, we qualitatively analyze the post-edited results of our model as shown in Table~\ref{tab:qualitative analysis_main}}.
A typical example of the multi-attribute correction appears in the first row, which changes the sentence structure and the preferred word choices. 
Our model tends to retain the overall meaning of the source sentence while precisely treating an abstractive personal \jh{behavior}. The output of PePe in the second row tends to keep loanwords in English instead of translating them into Korean (\textit{i.e., ``multi instrument'', ``preset load''}),
while \textit{mt} suffers from generating sentences that consider those preferences.
An example of changing a homonym to a suitable word is shown in the last row. Since \textit{``official building''} and \textit{``article''} are homonyms in Korean, PePe chooses the word that is appropriate for the semantic meaning of the sentence. 
We further provide several examples that consider the multidimensional attributes in Appendix~\ref{app: additional examples}. In either a single attribute or a multi attributes case, our model properly reflects distinct preferences.

\vspace{-0.1cm}

\section{Conclusion}
In this work, we propose a personalized post-editing method, PePe, utilizing user-generated post-edits.
Based on the APE framework, PePe leverages two modules, 1) user-specific parameters and 2) a semantic cluster-based discriminator module. 
These modules lead to reflect the multifarious inter-personal variations, where the former allows the model to learn user-dependent probabilities for each word while the latter unlearns the detrimental features in a pre-trained language model and maintains advantageous effects of the transfer learning.
We empirically demonstrate that PePe reflects fine-grained user preference in a variety of settings. 
To the best of our knowledge, this work is the first attempt to utilize the APE framework with the user-generated post-edits for personalized translation.
We believe that our work can draw more attention toward personalized translation, which is the ultimate direction that the neural translation model should go forward.

\bibliography{custom}
\bibliographystyle{acl_natbib}

\appendix
\section*{Supplementary Material}

This material complements our paper with additional experimental results and miscellaneous details.
Section~\ref{app: training details} provides the implementation details.
Section~\ref{app: robustness and hyperparameter} addresses the additional experiments that show the robustness of our model against a varied number of clusters and adjustment rates.
In Section~\ref{app: additional examples}, we demonstrate the variety of qualitative examples of post-edited sentences generated from PePe.

\section{Training Details}
\label{app: training details}

\paragraph{\textbf{Data Prepossessing.}} For the data preprocessing, we first filter out the duplicate lines and normalize the data such that each line represents a single sentence. Also, we exclude sentence that is longer than 100 words. Then, we utilize term frequency inverse document frequency (TF-IDF) to compute the user similarity score and filter out the noisy users. To be specific, we form a document for each user by aggregating \textit{src}. If a particular user has a lower than 0.1 similarity score, we exclude those users. We assume that if a user has a lower similarity score with others, then those users may contain noisy sentences. After prepossessing noisy data for USP, we divide the dataset into train/valid/test, which results in 5,207, 1,001, and 1,125 samples for Korean to English language pair, and 6,330, 1,360, and 1,357 samples for English to Korean. Since we split into train/valid/test for each user, the user appearing in the train dataset guarantees to appear in the test dataset. 

\paragraph{\textbf{Training and Inference Procedures.}} 
The main difference between training and inference procedures is the existence of a discriminator module. In other words, the clustering module and the discriminator are not utilized during the inference procedure. However, similar to the training procedure, we utilize the trained user-specific bias vector that corresponds to the user ID of each input sentence while generating a post-edited sentence.

\paragraph{\textbf{Evaluation and HyperParameter Details.}}We evaluate all experiments based on SacreBLEU\footnote{https://github.com/mjpost/sacrebleu}, TER\footnote{https://www.statmt.org/wmt18/ape-task.html}, and YiSi-1\footnote{https://github.com/chikiulo/yisi} scores. Since YiSi-1 requires pre-trained word embedding vectors, we utilize fastText\footnote{https://github.com/facebookresearch/fastText} to pretrain word embeddings. 
For the hyper-parameter settings, we use 10 clusters with 0.3 adjustment rates for all the experiments in the main paper. We select the combination of hyperparameters by manual tuning, which achieves the highest performance in the validation set based on the TER metric.
Conditions for early-stopping and decoding are equally applied to the baselines.
We follow the settings of hyperparameters in \citet{correia-martins-2019-simple} except sharing the weight of the encoder and the decoder.
We conducted all the experiments five times, and the random seeds used were 42, 1215, 101, 909, and 1129.
We selected the highest performance learning rate value between 0.00005 and 0.0001.
We report the configuration of our best model in Table~\ref{tab:confic}. 
\paragraph{\textbf{Environment Details.}}
All experiments in Table~\ref{tab:main_result} is examined with CentOS Linux release 7.8.2003, Tesla P40 GPU, and Intel Xeon CPU E5-2630. 
Results in Table~\ref{tab:ted} are examined with Ubuntu 16.04.6, Intel Xeon processor, and Tesla V100-PCIE-32GB GPU. The versions of the libraries we used in all experiments are 3.7.6 for Python and 1.4.0 for Pytorch. 

\begin{table}
\centering
\begin{tabular}{l|c}
\toprule
Hyperparameters & Value\\
\midrule
Pre-trained LM      & BERT-base-multilingual\\   
Learning rate    & 0.00005\\
Batch size       & 512\\
Accumulation step     & 2\\
Optimizer     & AdamW\\
Dropout     & 0.1\\
Label smoothing     & 0.1\\
Random seed & 42, 101, 1215, 1129, 909 \\
Decoding strategy & Beam search \\
Beam size & 3 \\

\bottomrule
\end{tabular}
\caption{Hyperparameter settings. AdamW~\citep{adamw} is the Adam~\cite{adam} optimizer with weight decay.}
\label{tab:confic}
\end{table}

\begin{table}[]
\centering
\begin{tabular}{lccc}
\toprule
Models & BLEU$\uparrow$ & TER$\downarrow$  \\
\midrule
Uncorrected  & 64.9          & 21.1 \\
DS-BERT  & 68.5          & 21.1 \\
\midrule
\textbf{PePe (k30, m0.1)}      & \textbf{70.2} & 20.2 \\   
PePe (k30, m0.2)    & 69.7          & 20.3 \\
\textbf{PePe (k30, m0.3)}       & 69.9          & \textbf{19.9} \\
PePe (k30, m0.4)     & 69.0          & 20.8 \\
PePe (k30, m0.5)     & 70.2          & 20.3 \\
\bottomrule
\end{tabular}
\caption{Experiments on various hyperparameter settings on a USP dataset. $k$ denotes the number of clusters and $m$ denotes the adjustment rate.}
\label{tab:hyperparameter}
\end{table}

\section{Robustness to the number of cluster and hyperparameter}
\label{app: robustness and hyperparameter}

In the main paper, we conduct all experiments with 10 cluster labels.
However, to be useful for the varied settings, it is crucial to demonstrate the model's robustness to the number of clusters and adjustment rate.
Here we provide the results trained on 30 cluster labels with various adjustment rates from 0.1 to 0.5. Identical with Table~\ref{tab:main_result}, we utilize \textit{en$\rightarrow$ko} dataset of 30 users.
Table~\ref{tab:hyperparameter} indicates that PePe consistently generates high-quality sentences, regardless of the hyperparameters. 

\section{Additional Qualitative Examples}
\label{app: additional examples}

This section provides additional qualitative examples from PePe. 
We choose the samples from the inference results of USP dataset, and
both \textit{ko$\rightarrow$en} and \textit{en$\rightarrow$ko} language pairs are reported.
As shown in Table~\ref{tab:qualitative analysis_add}, Table~\ref{tab:qualitative analysis_add2}, Table~\ref{tab:qualitative analysis_add3}, Table~\ref{tab:qualitative analysis_add4}, Table~\ref{tab:qualitative analysis_add5}, and
Table~\ref{tab:qualitative analysis_add6},
the tables are organized according to the typical personalization cases (\textit{i.e.,} error correction, word choice, politeness, and multiple attributes).
\textcolor{myr}{Red color} represents error correction case, \textcolor{myo}{Yellow color} represents word choice case, and \textcolor{myg}{Green color} represents politeness case. 
Each case also accompanies the insertion and deletion of the words (\textit{i.e.,} tokens).
Sentences inferred from PePe show that it well reflects the personal traits of each user and the characteristics of each language.

\begin{table*}
\footnotesize
\centering
\begin{tabular}{ll}
\toprule
\multicolumn{2}{l}{Error Correction (\textit{en$\rightarrow$ko})}\\
\midrule
\textit{src} & Begin the \underline{stroke} by moving the hand , while the elbow remains still and high.\\
\textit{mt} & 팔꿈치가 가만히 있고 높게 유지되는 동안 손을 움직이면서 \textcolor{myr}{뇌졸중}을 시작한다.\\
\textbf{PePe} & 팔꿈치가 가만히 있고 높게 유지되는 동안 손을 움직이면서 \textbf{\textcolor{myr}{팔동작}}을 시작한다.\\
\textit{pe} & 팔꿈치가 가만히 있고 높게 유지되는 동안 손을 움직이면서 \textbf{\textcolor{myr}{팔동작}}을 시작한다.\\
\midrule
\textit{src} & Periodically check on \underline{her} progress.\\
\textit{mt} & 정기적으로 진행 상황을 확인합니다.\\
\textbf{PePe} & 정기적으로 \textbf{\textcolor{myr}{그녀의}} 진행 상황을 확인합니다.\\
\textit{pe} & 정기적으로 \textbf{\textcolor{myr}{그녀의}} 진행 상황을 확인합니다.\\
\midrule
src & All \underline{manual checks} unclaimed for more than 6 months shall be canceled.\\
mt & 6개월 이상 청구되지 않은 모든 \textcolor{myr}{수동 점검}은 취소된다.\\
\textbf{PePe} & 6개월 이상 청구되지 않은 모든 \textbf{\textcolor{myr}{수동 수표}}는 취소된다.\\
pe & 6개월 이상 청구되지 않은 모든 \textbf{\textcolor{myr}{수동 수표}}는 취소된다.\\
\midrule
\textit{src} & If a signal has finite \underline{power} its energy will be infinite.\\
\textit{mt} & 만약 어떤 신호가 한정된 \textcolor{myr}{힘}을 가지고 있다면 그 에너지는 무한할 것이다.\\
\textbf{PePe} & 만약 어떤 신호가 한정된 \textbf{\textcolor{myr}{전력}}을 가지고 있다면 그 에너지는 무한할 것이다.\\
\textit{pe} & 만약 어떤 신호가 한정된 \textbf{\textcolor{myr}{전력}}을 가지고 있다면 그 에너지는 무한할 것이다.\\
\midrule
\textit{src} & The \underline{historical cost} of the intangible fixed assets transferred shall be the \underline{historical cost} recorded in the accounting \\ & records of the receiver.\\
\textit{mt} & 이전하는 무형고정자산의 \textcolor{myr}{역사적원가}는 수취인의 회계기록에 기록된 \textcolor{myr}{역사적원가}를 말한다.\\
\textbf{PePe} & 이전하는 무형고정자산의 \textbf{\textcolor{myr}{취득원가}}는 수취인의 회계기록에 기록된 \textbf{\textcolor{myr}{취득원가}}를 말한다.\\
\textit{pe} & 이전하는 무형고정자산의 \textbf{\textcolor{myr}{취득원가}}는 수취인의 회계기록에 기록된 \textbf{\textcolor{myr}{취득원가}}를 말한다.\\
\midrule
\textit{src} & \underline{Emotional exhaustion} is the central quality and the most obvious manifestation of burnout.\\
\textit{mt} & \textcolor{myr}{감정 기진맥진}은 중심적인 질이고 가장 명백한 소진 증상이다.\\
\textbf{PePe} & \textbf{\textcolor{myr}{정서적 소진}}은 중심적인 질이고 가장 명백한 번아웃 증상이다.\\
\textit{pe} & \textbf{\textcolor{myr}{정서적 소진}}은 번아웃의 중심특성이자 가장 명백한 징후이다.\\
\bottomrule
\end{tabular}
\caption{PePe generates post-edited sentences that corrects the grammar errors from the machine-translated outputs.}
\label{tab:qualitative analysis_add}
\end{table*}

\begin{table*}[!htbp]
\footnotesize
\centering
\begin{tabular}{ll}
\toprule
\multicolumn{2}{l}{Word Choice (\textit{en$\rightarrow$ko})}\\
\midrule
\textit{src} & Hide the layer containing the \underline{cutting lines}.\\
\textit{mt} & \textcolor{myo}{절단선}이 들어 있는 레이어를 숨긴다.\\
\textbf{PePe} & \textbf{\textcolor{myo}{커팅 라인}}이 들어 있는 레이어를 숨긴다.\\
\textit{pe} & \textbf{\textcolor{myo}{커팅 라인}}이 들어 있는 레이어를 숨긴다.\\
\midrule
\textit{src} & \underline{The worker} explores cultural diversity factors that may be a part of the problem or situation.\\
\textit{mt} & \textcolor{myo}{노동자}는 문제나 상황의 일부일 수도 있는 문화적 다양성 요소를 탐구한다.\\
\textbf{PePe} & \textbf{\textcolor{myo}{사회복지사}}는 문제나 상황의 일부일 수도 있는 문화적 다양성 요소를 탐구한다.\\
\textit{pe} & \textbf{\textcolor{myo}{사회복지사}}는 문제나 상황의 일부일 수도 있는 문화적 다양성 요소를 탐구한다.\\
\midrule
\textit{src} & Exemestane is one of the most potent \underline{aromatase} inhibitors presently available.\\
\textit{mt} & exemestane은 현재 사용 가능한 가장 강력한 \textcolor{myo}{방향족} 억제제 중 하나이다.\\
\textbf{PePe} & exemestane은 현재 사용 가능한 가장 강력한 \textbf{\textcolor{myo}{aromatase}} 억제제 중 하나이다.\\
\textit{pe} & exemestane은 현재 사용 가능한 가장 강력한 \textbf{\textcolor{myo}{aromatase}} 억제제 중 하나이다.\\
\midrule
\textit{src} & \underline{You} do not want them drunk and lazy.\\
\textit{mt} & \textcolor{myo}{너는} 그들이 술에 취해서 게을러지는 것을 원하지 않는다.\\
\textbf{PePe} & \textbf{\textcolor{myo}{당신은}} 그들이 술에 취해서 게을러지는 것을 원하지 않는다.\\
\textit{pe} & \textbf{\textcolor{myo}{당신은}} 그들이 술에 취해서 게을러지는 것을 원하지 않는다.\\
\midrule
\textit{src} & By combining the two outputs without the external phase shift, a sum signal is provided for \underline{range} tracking.\\
\textit{mt} & 외부 위상 이동 없이 두 출력을 결합함으로써 \textcolor{myo}{범위} 추적을 위한 합계 신호가 제공된다.\\
\textbf{PePe} & 외부 위상 이동 없이 두 출력을 결합함으로써 \textbf{\textcolor{myo}{거리}} 추적을 위한 합계 신호가 제공된다.\\
\textit{pe} & 외부 위상 이동 없이 두 출력을 결합함으로써 \textbf{\textcolor{myo}{거리}} 추적을 위한 합계 신호가 제공된다.\\
\midrule
\textit{src} & What are my \underline{needs} for developing my capacity and potentiality?\\
\textit{mt} & 내 능력과 잠재력을 개발하기 위한 나의 \textcolor{myo}{필요성은} 무엇인가?\\
\textbf{PePe} & 내 능력과 잠재력을 개발하기 위한 나의 \textbf{\textcolor{myo}{needs는}} 무엇인가?\\
\textit{pe} & 내 능력과 잠재력을 개발하기 위한 나의 \textbf{\textcolor{myo}{needs는}} 무엇인가?\\
\bottomrule
\end{tabular}
\caption{PePe changes the words that are not suitable for personal style but are grammatically correct to other candidates, such as synonyms and loanwords.
}
\label{tab:qualitative analysis_add2}
\end{table*}

\begin{table*}[!htbp]
\footnotesize
\centering
\begin{tabular}{ll}
\toprule
\multicolumn{2}{l}{Politeness (\textit{en$\rightarrow$ko})}\\
\midrule
\textit{src} & Spend some time looking over the meeting agenda in advance and think about some of the key topics. \\
\textit{mt} & 회의 안건을 미리 살펴보고 몇 가지 주요 주제에 대해 \textcolor{myg}{생각해 보십시오}.  \\
\textbf{PePe} & 회의 안건을 미리 살펴보고 몇 가지 주요 주제에 대해 \textbf{\textcolor{myg}{생각해 보라}}. \\
\textit{pe} & 회의 안건을 미리 살펴보고, 몇 가지 주요 주제에 대해 \textbf{\textcolor{myg}{생각해 보라}}.  \\
\midrule
\textit{src} & She wants the assignment. \\
\textit{mt} & 그녀는 그 과제를 \textcolor{myg}{원한다}.\\
\textbf{PePe} & 그녀는 그 과제를 \textbf{\textcolor{myg}{원합니다}}.\\
\textit{pe} & 그녀는 그 과제를 \textbf{\textcolor{myg}{원합니다}}.\\
\midrule
\textit{src} & \underline{Watch} this video for directions on how to complete the S1 Conversations challenge.\\
\textit{mt} & 대화 과제를 완료하는 방법은 이 비디오를 \textcolor{myg}{참조하십시오}.\\
\textbf{PePe} & 대화 과제를 완료하는 방법은 이 비디오를 \textbf{\textcolor{myg}{참조하세요}}.\\
\textit{pe} & 대화 과제를 완료하는 방법은 이 비디오를 \textbf{\textcolor{myg}{참조하세요}}.\\
\midrule
\textit{src} & According to the U.S. Bureau of Census, there are approximately 90 million households in the United States.\\
\textit{mt} & 미국 인구조사국에 따르면, 미국에는 약 9천만 가구가 살고 있다고 \textcolor{myg}{합니다}.\\
\textbf{PePe} & 미국 인구조사국에 따르면, 미국에는 약 9천만 가구가 살고 있다고 \textbf{\textcolor{myg}{한다}}.\\
\textit{pe} & 미국 인구 조사국에 따르면, 미국에는 약 9천만 가정이 살고 있다고 \textbf{\textcolor{myg}{한다}}.\\
\midrule
\textit{src} & The store is located inside the Terminal 1.\\
\textit{mt} & 그 상점은 터미널 1 안에 \textcolor{myg}{있다}.\\
\textbf{PePe} & 지점은 터미널 1 안에 \textbf{\textcolor{myg}{있습니다}}.\\
\textit{pe} & 지점은 터미널 1 내에 \textbf{\textcolor{myg}{있습니다}}.\\
\midrule
\textit{src} & Maps are also available that show the tract boundaries, making the data readily discernible.\\
\textit{mt} & 트랙 경계가 표시된 지도도 사용할 수 있어 데이터를 쉽게 식별할 수 \textcolor{myg}{있습니다}.\\
\textbf{PePe} & 트랙 경계가 표시된 지도도 사용할 수 있어 데이터를 쉽게 식별할 수 \textbf{\textcolor{myg}{있다}}.\\
\textit{pe} & 통로 경계를 보여주는지도도 제공되므로 데이터를 쉽게 식별 할 수 \textbf{\textcolor{myg}{있다}}.\\
\bottomrule
\end{tabular}
\caption{PePe controls the level of politeness. The usage of the honorifics varies from language to language. 
}
\label{tab:qualitative analysis_add3}
\end{table*}

\begin{table*}[!htbp]
\footnotesize
\centering
\begin{tabular}{ll}
\toprule
\multicolumn{2}{l}{Multiple Attributes (\textit{en$\rightarrow$ko})}\\
\midrule
\textit{src} & Our staff will send you back to the airport.\\
\textit{mt} & \textcolor{myg}{우리} 직원이 너를 공항으로 돌려보낼 \textcolor{myg}{것이다}.\\
\textbf{PePe} & \textbf{\textcolor{myg}{저희}} 직원이 공항으로 돌려보낼 \textbf{\textcolor{myg}{것입니다}}.\\
\textit{pe} & \textbf{\textcolor{myg}{저희}} 직원이 고객님을 공항으로 데려다 줄 \textbf{\textcolor{myg}{것입니다}}.\\
\midrule
\textit{src} & When the \underline{machine} receives the data, it automatically reads the \underline{crop marks} using a sensor, and then starts \textbf{\textcolor{myo}{cutting}}.\\
\textit{mt} & 
\textcolor{myo}{기계}가 데이터를 받으면 자동으로 센서를 이용해 \textcolor{myo}{자르기 표시}를 읽은 뒤 \textcolor{myo}{절단}을 시작한다.\\
\textbf{PePe} & \textbf{\textcolor{myo}{장비}}가 데이터를 받으면 자동으로 센서를 이용해 \textbf{\textcolor{myo}{crop mark}}를 읽은 뒤 \textbf{\textcolor{myo}{커팅}}을 시작한다.\\
\textit{pe} & \textbf{\textcolor{myo}{장비}}가 데이터를 받으면 자동으로 센서를 이용해 \textbf{\textcolor{myo}{crop mark}}를 읽은 뒤 \textbf{\textcolor{myo}{커팅}}을 시작한다.\\
\midrule
\textit{src} & When transferring major repairs of fixed assets for \underline{non-business activities}, \underline{the following accounts shall be recorded}.\\
\textit{mt} & \textcolor{myo}{비사업활동용} 고정자산의 주요수리를 이전할 때에는 \textcolor{myg}{다음 사항을 기재하여야 한다}.\\
\textbf{PePe} & \textbf{\textcolor{myo}{비영리활동용}} 고정자산의 주요수리를 이전할 때에는 \textbf{\textcolor{myr}{다음과 같이 회계처리 하여야 한다}}.\\
\textit{pe} & \textbf{\textcolor{myo}{비영리활동용}} 고정자산의 주요수리를 이전할 때에는 \textbf{\textcolor{myr}{다음과 같이 회계처리 하여야 한다}}.\\
\midrule
\textit{src} & The free shuttle bus will come to pick \underline{you} up around 10 minutes.\\
\textit{mt} & 무료 셔틀버스가 약 10분 정도 \textcolor{myo}{당신}을 데리러 올 \textcolor{myg}{것이다}.\\
\textbf{PePe} & 무료 셔틀버스가 약 10분 정도 \textbf{\textcolor{myo}{고객님}}을 데리러 올 \textbf{\textcolor{myg}{것입니다}}.\\
\textit{pe} & 무료 셔틀버스가 약 10분 정도 \textbf{\textcolor{myo}{고객님}}을 데리러 올 \textbf{\textcolor{myg}{것입니다}}.\\
\midrule
\textit{src} & If using 3 \underline{crop marks}, select 3-point start.\\
\textit{mt} & 3개의 \textcolor{myo}{자르기 표시}를 사용하는 경우 3-point start를 \textcolor{myg}{선택하십시오}.\\
\textbf{PePe} & 3개의 \textbf{\textcolor{myo}{crop mark}}를 사용하는 경우 3-point start를 \textbf{\textcolor{myg}{선택한다}}.\\
\textit{pe} & 3개의 \textbf{\textcolor{myo}{crop mark}}를 사용하는 경우 3-point start를 \textbf{\textcolor{myg}{선택한다}}.\\
\midrule
\textit{src} & The following \underline{parameters} control the display of \underline{points-clouds} (right).\\
\textit{mt} & 다음 \textcolor{myo}{매개 변수}는 \textcolor{myr}{점 구름} (오른쪽) 의 표시를 제어합니다.\\
\textbf{PePe} & 다음 \textbf{\textcolor{myo}{파라미터}}는 \textbf{\textcolor{myr}{포인트 클라우드}} (오른쪽) 의 표시를 제어합니다.\\
\textit{pe} & 다음 \textbf{\textcolor{myo}{파라미터}}는 \textbf{\textcolor{myr}{포인트 클라우드}} (오른쪽) 의 표시를 제어합니다.\\
\bottomrule
\end{tabular}
\caption{PePe not only tackles a single attribute but also generates high-quality sentences with multiple attributes revised. Each attribute is colored with a corresponding color.}
\label{tab:qualitative analysis_add4}
\end{table*}

\begin{table*}[!htbp]
\small
\centering
\begin{tabular}{ll}
\toprule
\multicolumn{2}{l}{Error Correction (\textit{ko$\rightarrow$en})}\\
\midrule
\textit{src} & 그래서 \underline{전치사} `reo'는 `to'와 `for'의 의미가 \underline{있다}.\\
\textit{mt} & so the \textcolor{myr}{prepositions} `reo' \textcolor{myr}{have} the meaning of `to' and `for'.\\
\textbf{PePe} & so the \textbf{\textcolor{myr}{preposition}} `reo' \textbf{\textcolor{myr}{has}} the meaning of `to' and `for'.\\
\textit{pe} & so the \textbf{\textcolor{myr}{preposition}} `reo' \textbf{\textcolor{myr}{has}} the meaning of `to' and `for'.\\
\midrule
\textit{src} & \underline{관사} 아래에 있는 모음코드가 이렇게 바뀌어진다.\\
\textit{mt} & the vowel code under the \textcolor{myr}{official building} changes like this.\\
\textbf{PePe} & the vowel code under the \textbf{\textcolor{myr}{article}} changes like this.\\
\textit{pe} & the vowel code under the \textbf{\textcolor{myr}{article}} changes like this.\\
\midrule
\textit{src} & 
\underline{professor} gasser는 단일한 관리통제 기구보다 상업적 차원, 정부 차원 등 다차원적 모델의 시도를 결합하는 노력이 \\
& 필요하다고 지적한다.\\
\textit{mt} & the \textcolor{myr}{processor} gasser points out that efforts need to be made to combine attempts by multi-dimensional models such as \\
& commercial and government levels rather than single management and control organizations.\\
\textbf{PePe} & the \textbf{\textcolor{myr}{professor}} gasser points out that efforts need to be made to combine attempts by multi-dimensional models such as \\
& commercial and government levels rather than single management and control organizations.\\
\textit{pe} & the \textbf{\textcolor{myr}{professor}} gasser points out that efforts need to be made to combine attempts by multi-dimensional models such as \\
& commercial and government levels rather than single management and control organizations.\\
\midrule
\textit{src} & `dagesh'가 \underline{놓일} 수 없다.\\
\textit{mt} & `dagesh' can't be \textcolor{myr}{let go}.\\
\textbf{PePe} & `dagesh' can't be \textbf{\textcolor{myr}{placed}}.\\
\textit{pe} & `dagesh' can't be \textbf{\textcolor{myr}{placed}}.\\
\bottomrule
\end{tabular}
\caption{PePe generates post-edited sentences that corrects the grammar errors from the machine-translated outputs.}
\label{tab:qualitative analysis_add5}
\end{table*}

\begin{table*}[!htbp]
\small
\centering
\begin{tabular}{ll}
\toprule
\multicolumn{2}{l}{Word Choice (\textit{ko$\rightarrow$en})}\\
\midrule
\textit{src} & 내부배선의 \underline{색상}은 아래와 같이 구분하여 사용하여야 한다.\\
\textit{mt} & the \textcolor{myo}{colour} of the inner wiring shall be used separately as follows.\\
\textbf{PePe} & the \textbf{\textcolor{myo}{color}} of the inner wiring shall be used separately as follows.\\
\textit{pe} & the \textbf{\textcolor{myo}{color}} of the inner wiring shall be used separately as follows.\\
\midrule
\textit{src} & 추정 공시가격이 올해 거래된 \underline{urgent} sale price를 앞서고 있다.\\
\textit{mt} &  the estimated official price is ahead of the \textcolor{myo}{current} sales price traded this year.\\
\textbf{PePe} & the estimated official price is ahead of the \textbf{\textcolor{myo}{urgent}} sale price traded this year.\\
\textit{pe} & the estimated official price is ahead of the \textbf{\textcolor{myo}{urgent}} sales price traded this year.\\
\midrule
\textit{src} & 12월 싱가폴,말레이시아지역 \underline{패키지} 상품 판매 확대\\
\textit{mt} & expanding sales of \textcolor{myo}{package} products in singapore and malaysia in december.\\
\textbf{PePe} & expanding sales of \textbf{\textcolor{myo}{pkg}} products in singapore and malaysia in december.\\
\textit{pe} & expanding sales of \textbf{\textcolor{myo}{pkg}} products in singapore and malaysia in december.\\
\midrule
\textit{src} & 한해 전에 쓰고 남은 돈이 1억2천만\underline{원} 정도였다.\\
\textit{mt} & the remaining money was about 120 million \textcolor{myo}{won} a year ago.\\
\textbf{PePe} & the remaining money was about \textbf{\textcolor{myo}{krw}} 120 million a year ago.\\
\textit{pe} & the remaining money was about \textbf{\textcolor{myo}{krw}} 120 million a year ago.\\
\bottomrule
\end{tabular}
\caption{PePe changes the words that are not suitable for personal style but are grammatically correct to other candidates, such as synonyms and loanwords.}
\label{tab:qualitative analysis_add6}
\end{table*}

\end{document}